\documentclass[sigconf]{acmart}
\AtBeginDocument{%
  }

\begin{document}

\title{\textit{StayLTC}: A Cost-Effective Multimodal Framework for Hospital Length of Stay Forecasting}




\author{Sudeshna Jana}
\affiliation{%
 \institution{TCS Research}
 \city{Kolkata}
 \state{West Bengal}
 \country{India}}
\email{sudeshna.jana@tcs.com}

\author{Manjira Sinha}
\affiliation{%
 \institution{TCS Research}
 \city{Kolkata}
 \state{West Bengal}
 \country{India}}
\email{sinha.manjira@tcs.com}

\author{Tirthankar Dasgupta}
\affiliation{%
 \institution{TCS Research}
 \city{Kolkata}
 \state{West Bengal}
 \country{India}}
\email{dasgupta.tirthankar@tcs.com}




\begin{abstract}
  Accurate prediction of Length of Stay (LOS) in hospitals is crucial for improving healthcare services, resource management, and cost efficiency. This paper presents StayLTC, a multimodal deep learning framework developed to forecast real-time hospital LOS using Liquid Time-Constant Networks (LTCs). LTCs, with their continuous-time recurrent dynamics, are evaluated against traditional models using structured data from Electronic Health Records (EHRs) and clinical notes. Our evaluation, conducted on the MIMIC-III dataset, demonstrated that LTCs significantly outperform most of the other time series models, offering enhanced accuracy, robustness, and efficiency in resource utilization. Additionally, LTCs demonstrate a comparable performance in LOS prediction compared to time series large language models, while requiring significantly less computational power and memory, underscoring their potential to advance Natural Language Processing (NLP) tasks in healthcare.
\end{abstract}

\ccsdesc[500]{Information systems~Information retrieval}
\ccsdesc[500]{Information systems~Retrieval tasks and goals}
\ccsdesc[500]{Information systems~Information extraction}

\keywords{Hospital Length of Stay, Liquid Neural Networks
Memory Efficiency, MIMIC Dataset, Clinical Notes}


\maketitle

\section{Introduction}
The growing demand for healthcare services, driven by aging populations, has made healthcare expenditure a major part of GDP \cite{mohapatra2024establishing}. As a result, cost containment is a key challenge for healthcare management today.\cite{veena2023applications} Hospital Length of Stay (LOS), the days a patient spends in a facility per admission, is a crucial measure of resource use. Accurate LOS predictions help optimize bed management, staffing, and critical equipment usage, reducing costs and improving discharge planning. \cite{diwan2020effect, nelson2022mortality} This also aids policymakers in budgeting and aligning financial goals with patient outcomes.\cite{tabish2024healthcare}

Electronic Health Records (EHRs) are essential for developing predictive models for hospital LOS, using both structured data (e.g., demographics, lab results) and unstructured data (e.g., clinical notes). Researchers have applied various machine learning models to predict LOS. For instance, \cite{cai2016real} used a Bayesian Network for real-time LOS predictions, and \cite{chrusciel2021prediction} improved predictions by incorporating clinical notes. Recent studies, like \cite{zeleke2023machine}, compared machine learning models, focusing on patients' demographics and clinical information to predict LOS more accurately. While classification techniques are commonly used to predict LOS, \cite{alsinglawi2020predicting, zeleke2024comparison} employed various machine learning regression models to forecast the actual number of LOS days.

Most prior studies have focused on structured EHR data, like demographics, vital signs, and lab results. However, clinical notes in EHRs offer valuable insights into patient conditions, treatments, diagnoses and adverse effects. Advances in Natural Language Processing (NLP) and Large Language Models (LLMs) now enable better analysis of these unstructured notes. As a result, researchers are increasingly integrating unstructured data into predictive models. Predicting exact LOS is rare; instead, predictions are usually grouped into short, medium, or long stays.

In this paper, we propose a novel multimodal time series prediction framework, \textit{StayLTC}, to forecast the remaining hospital length of stay for patients at the beginning of each day post-admission. The model leverages Liquid Time-Constant Networks (LTCs), \cite{hasani2021liquid} known for their continuous-time recurrent dynamics, and integrates time-stamped clinical notes with structured vital parameters. Moreover, unlike transformer-based models and large language models, LTCs require significantly fewer parameters, making them much more lightweight, domain-agnostic and resource-efficient. Tested on the MIMIC-III database,\cite{johnson2016mimic} LTCs offer efficient parameter training, requiring far fewer resources than transformer models and LLMs. This is the first study to apply LTCs to LOS prediction, highlighting their efficiency compared to other time series models. 

\begin{figure*}
  \centering
  \includegraphics[width=.8\linewidth]{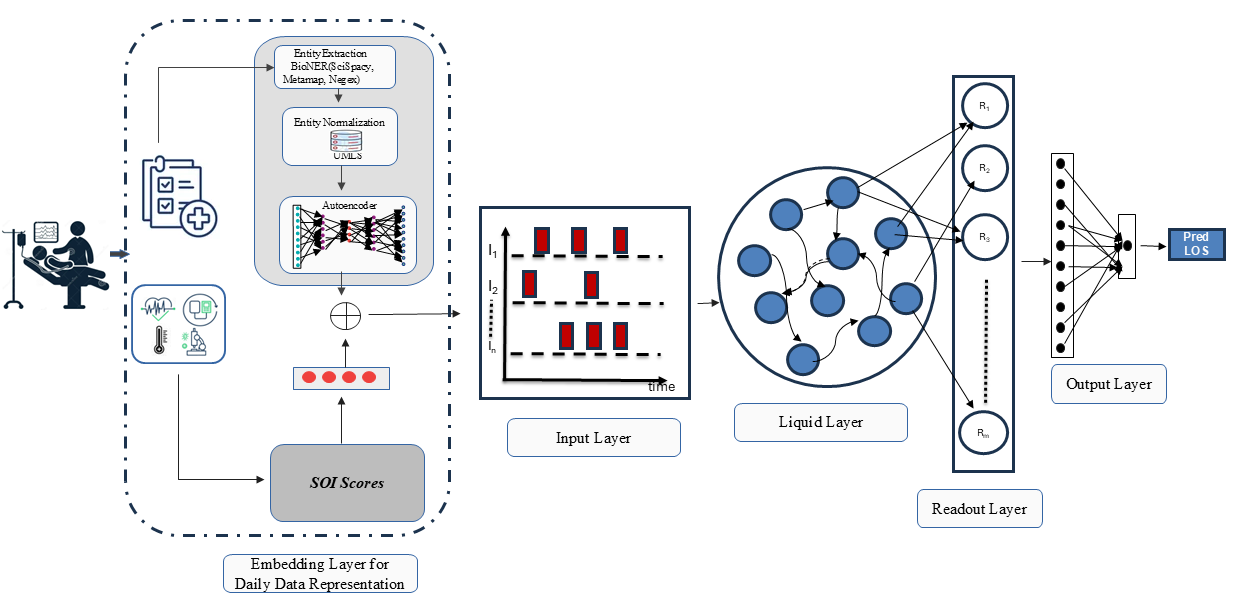}
  \caption{Schematic overview of the \textit{StayLTC} framework for predicting LOS.}
  \label{StayLTC}
\end{figure*}

\section{Proposed \textit{StayLTC} Framework}\label{method}
This section outlines our multimodal framework (Figure \ref{StayLTC}) for predicting the remaining length of stay (LOS) in the hospital at the start of each day, based on the patient's health progression or deterioration. The framework integrates daily clinical reports (e.g., nursing, radiology, ECG) along with vital signs and laboratory measurements.

\subsection{Processing of Unstructured Clinical Notes}
Clinical notes exhibit considerable variability in both style and content. Some notes solely document symptoms, while others also include mentions of the absence of symptoms, adverse reactions, psychological states, and changes in appetite, often using non-standard terminology and abbreviations. To address this variability, we implemented a processing layer that leverages biomedical dictionaries to create a structured representation of the clinical details. This includes extracting clinical entities such as \textit{diseases or symptoms}, \textit{injuries or poisoning}, \textit{abnormalities}, \textit{lifestyle}, \textit{mental health conditions}, and \textit{previous health histories}, using two BioNER tools: ScispaCy specifically, \textit{en\_ner\_bc5cdr\_md} model \cite{neumann2019scispacy} and Metamap \cite{aronson2001effective}. In addition to identifying entities, we employed the Negex algorithm \cite{mehrabi2015deepen} to detect absence indicators commonly found in clinical notes, such as ``absence of pain" or ``no history of hypertension". Moreover, clinical data often contains diverse non-standard terminology, abbreviations, various formats, and coding systems to represent similar clinical concepts. For example, ``Pulmonary Edema" and ``fluid in lungs" refer to the same condition. To standardize these entities, we utilized the UMLS Metathesaurus \cite{schuyler1993umls} API, which assigns a unique "Concept Unique Identifier (CUI)" to each concept.

Once entities are extracted and represented with CUIs, each day’s clinical details for a patient are consolidated using the CUIs observed on that day. Given a patient $p$, the clinical details at day $t$ is defined by a vector $H_p(t) = <f(d_i)>$ , $i = 1, 2,..., |V|$  , where $d_i \in V$ and the value of $f(d_i)$ is set to 1 if $d_i$ present, -1 if it is mentioned negatively, and 0 if $d_i$ is not mentioned in day $t$.

Additionally, the high number of unique diseases and symptoms, along with the variability in individual manifestations, leads to high-dimensional and sparse vectors. To address this, we employed a standard autoencoder (AE) framework \cite{wang2016auto} to obtain a dense, lower-dimensional representation. The AE is an unsupervised model where the ``encoder'' network compresses the input data by capturing its key features, and the ``decoder'' network reconstructs the original data from this compressed form, aiming to preserve the essential information.

Let $X\in \mathbb{R}^{m\times n}$ represent the input data, where $n$ represents the number of samples and $m$ represents the number of features of each sample. The AE optimizes the following loss function to minimize the reconstruction error:
\[
\mathcal{L}(\mathbf{X}, \hat{\mathbf{X}}) = \frac{1}{N} \sum_{i=1}^{N} \left[ X_i - g(f(X_i)) \right]^2
\]

In our experiment, the encoder used a multi-layer neural network to map the input data to a low-dimensional latent space, while the decoder employed the inverse structure of the encoder. The mean squared error (MSE) was used as the loss function, and the Adam optimizer with a learning rate of $0.001$ was applied to ensure model convergence. The compressed representations $f(X_i)$ provide a more efficient and informative vector representation of the patient's health conditions for subsequent tasks.


\subsection{Compute Severity of Illness (SOI) Scores}
In our model, we also integrated structured health parameters like vital signs and lab measurements, represented by the Severity of Illness (SOI) score, a key metric for assessing patient severity. We used four SOI scores namely, \textit{Acute Physiology and Chronic Health Evaluation (APACHE-II)} \cite{knaus1985apache}, \textit{Simplified Acute Physiology (SAPS-II)}\cite{le1993new}, \textit{Sepsis-related Organ Failure Assessment (SOFA)}\cite{moreno1999use}, and \textit{Oxford Acute Severity of Illness Score (OASIS)} \cite{chen2018clinical}. These scores indicate disease severity, complexity, and organ system impairment, offering valuable insights into a patient’s clinical status. All scores are generated using publicly available tools.

\subsection{LTC network for real time LOS prediction}
Afterward, we have used the Liquid Time-Continuous (LTC) network for time-series prediction of hospital LOS for individual patients. Each patient's daily health condition is represented as an autoencoded embedding vector of clinical notes say, $EH_t$. This embedding is concatenated with the SOI scores say $<s^1_t, s^2_t, s^3_t, s^4_t>$ from the patient's clinical measurements, forming the input vector for the LTC network at time $t$.

Recently, in 2021, Hasani et al. \cite{hasani2021liquid} introduced a new class of recurrent neural networks with continuous-time hidden states defined by ordinary differential equations (ODEs), making them particularly effective for time series data. The hidden state of this network, $x(t)$,  is determined by the solution to a system of linear ODEs of the following form:
\begin{center}
\label{eq1}
    $\frac{dx(t)}{dt} = -\left[\frac{1}{\tau} + f(x(t), I(t), t, \theta)\right]x(t)$ \\
  $+f(x(t), I(t), t, \theta)A$
\end{center}
where, $t$ represents time, $I(t)$ is input at time $t$, $\tau$ is the time-constant and the neural network $f$ is parametrized by $\theta$ and $A$. This LTC network offers several features and benefits: In the above equation, the neural network $f$, dynamically influences both the evolution of the hidden state and serves as an input-dependent varying time-constant $ \tau_{sys}$, for the learning system.
\begin{equation}
    \tau_{sys} = \frac{\tau}{1 + \tau f(x(t), I(t), t, \theta)}
\end{equation}
This property allows each element of the hidden state to capture different temporal patterns from specific input features at each time step, making LTCs effective for processing complex, time-varying data. 
Additionally, the authors demonstrated that the time constant and state of LTC neurons are bounded to a finite range. This state stability ensures that the outputs of LTCs do not explode, even when their inputs tend toward infinity. Furthermore, they demonstrated that Liquid Time-Constant (LTC) networks are universal approximators,\cite{hasani2018liquid} meaning they can approximate any autonomous ODE with a finite number of neurons to any desired precision. By evaluating expressivity in terms of trajectory length, they showed that LTC networks outperform existing time series models.

To build our Liquid Neural Network, we use the Neural Circuit Policies (NCP) wiring with the AutoNCP configuration \cite{lechner2020neural}. The network consists of 28 sensory neurons and 1 motor neuron as the output. In the AutoNCP framework, the remaining neurons are automatically parameterized. The Liquid Layer output is processed through a fully connected layer with a linear activation function. We train the model using the Mean Squared Error (MSE) loss function and the Adam optimizer with a learning rate of 0.001 to ensure effective learning and convergence.

\section{Results and Discussions}\label{results}




\begin{table}[ht]
\footnotesize
\addtolength{\tabcolsep}{-0.4em}
\caption{Performance metrics of baseline models for LOS predictions.}
\begin{tabular}{|l|l|l|l|l|} 

\hline
Model & Input & $R^2$ & MAE & RMSE \\
\hline
LSTM & HealthVector($H_p(t)$) & 0.53 & 0.45 & 0.47\\
\hline
LSTM & Healthvector($H_p(t)$)+SOI & 0.55 & 0.44 & 0.47\\
\hline
LSTM & AutoencodedHealthVector & 0.55 & 0.41 & 0.43\\
\hline
LSTM & AutoencodedHealthVector+SOI & 0.56 & 0.41 & 0.41\\
\hline
ClinicalBERT+LSTM & Notes & 0.53 & 0.52 & 0.58 \\
\hline
ClinicalBERT+LSTM & Notes+SOI & 0.54 & 0.50 & 0.57\\
\hline
BlueBERT+LSTM & Notes & 0.53 & 0.51 & 0.57 \\
\hline
BlueBERT+LSTM & Notes + SOI & 0.55 & 0.46 & 0.53\\
\hline
Informer & HealthVector($H_p(t)$) & 0.68 & 0.28 & 0.32\\
\hline
Informer & HealthVector($H_p(t)$)+SOI & 0.68 & 0.27 & 0.30\\
\hline
Informer & AutoencodedHealthVector & 0.70 & 0.25 & 0.30\\
\hline
Informer & AutoencodedHealthVector+SOI & 0.71 & 0.25 & 0.28\\
\hline
TIME-LLM & Notes & 0.85 & 0.19 & 0.25\\
\hline
TIME-LLM & Notes+SOI & \textbf{0.87} & \textbf{0.17} & \textbf{0.21}\\
\hline
LTC & HealthVector($H_p(t)$) & 0.67 & 0.25 & 0.29\\
\hline
LTC & HealthVector($H_p(t)$)+SOI & 0.69 & 0.25 & 0.26 \\
\hline
LTC & AutoencodedHealthVector & 0.74 & 0.23 & 0.25 \\
\hline
LTC & AutoencodedHealthVector+SOI& \textbf{0.78} & \textbf{0.20} & \textbf{0.24} \\
\hline

\end{tabular}
  \label{results-table}

\end{table}

\subsection{Dataset} 
The study involved 5,000 patients from the MIMIC-III v1.4 database \cite{johnson2016mimic}, which is IRB-approved and accessible after completing the CITI ``Data or Specimens Only Research'' course. We excluded patients under 18 years of age and those with hospital stays shorter than two days, or with incomplete records, as sufficient contextual knowledge is essential for LOS prediction. We only included patients with a length of stay under one month to eliminate outliers. Among the cohort, 45\% had pneumonia, 31\% had sepsis, and 24\% had cardiovascular diseases. The average hospital stay was 10 days.

\subsection{Baseline Models} 
We experimented and compared the performance of various time series models to predict the remaining hospital length of stay for patients at the start of each day following admission. We assessed the StayLTC model's performance based on two criteria: (a)  its efficiency in terms of space and time complexity compared to baseline models and (b) the prediction accuracy of the architecture. For LOS prediction, we explored several input modalities, including clinical notes, extracted health conditions, autoencoded health vectors, and combinations of these with severity of illness (SOI) scores. Additionally, we compared transformer-based embeddings from ClinicalBERT and BlueBERT with our own representations.

We evaluated the Informer model \cite{zhou2021informer}, a transformer-based time-series model known for its efficiency in handling long-range dependencies and large-scale data, outperforming traditional recurrent neural networks (RNNs). Its key innovation is the ProbSparse self-attention mechanism, which sparsifies attention by focusing only on a subset of relevant tokens, reducing both time and space complexity. The encoder processes the input through multiple layers of ProbSparse self-attention and feed-forward networks to capture temporal patterns, while the decoder uses cross-attention and a fully connected layer to generate future predictions. For training, we adapted Mean Squared Error (MSE) loss function, the Adam optimizer, and a learning rate of 0.001. 

Additionally, we explored TIME-LLM \cite{jin2023time}, a reprogramming framework designed to adapt existing LLMs for time series forecasting, with the backbone LLMs frozen. TIME-LLM begins by tokenizing the input time series using a customized patch embedding layer. These patches are then passed through a reprogramming layer, which transforms the forecasting task into a language task. To activate the LLM’s capabilities, authors proposed to use Prompt-as-Prefix (PaP) strategy to complement patch reprogramming. Finally, the transformed time series patches from the LLM are projected to obtain the forecasts. For this experiment, we implemented TIME-LLM in the \textit{NeuralForecast} library and used Llama2-7B \cite{touvron2023llama} as the backbone model.

\subsection{Results} From the selected cohort of 5,000 patients, we extracted 45,312 time-stamped clinical notes, including nursing, ECG, and radiology reports. After preprocessing, we compiled a list of 8,700 unique diseases and symptoms and generated 500-dimensional auto-encoded health vectors for each day. These vectors, along with SOI scores, were integrated into the predictive time series models.

Table \ref{results-table} summarizes the overall performance of our experiments for the LOS prediction task. The baseline model incorporating clinical notes with SOI scores outperformed the notes-only baseline, highlighting the added predictive value of severity information. The proposed StayLTC model outperformed all baseline models, except TIME-LLM showing a statistically significant improvement. Additionally, the health vectors generated from our preprocessing methods significantly enhanced performance compared to embeddings from ClinicalBERT and BlueBERT. While ClinicalBERT and BlueBERT embeddings capture linguistic nuances like ``severe pain'' vs ``mild pain'', they do not consistently reflect the similarities or differences in medical terminology used in the notes. 

\begin{figure}[!h]
  \centering
  \includegraphics[width=\linewidth]{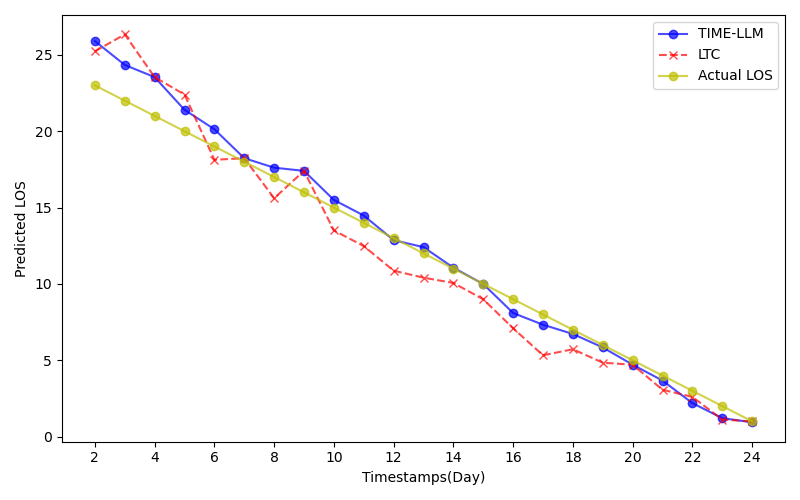}
  \caption{Predictions for a patient's real-time LOS using autoencoded health vectors and SOI scores: LTC vs. TIME-LLM}
  \label{LTC vs TIME-LLM}
\end{figure}

\begin{figure}[!h]
  \centering
  \includegraphics[width=\linewidth]{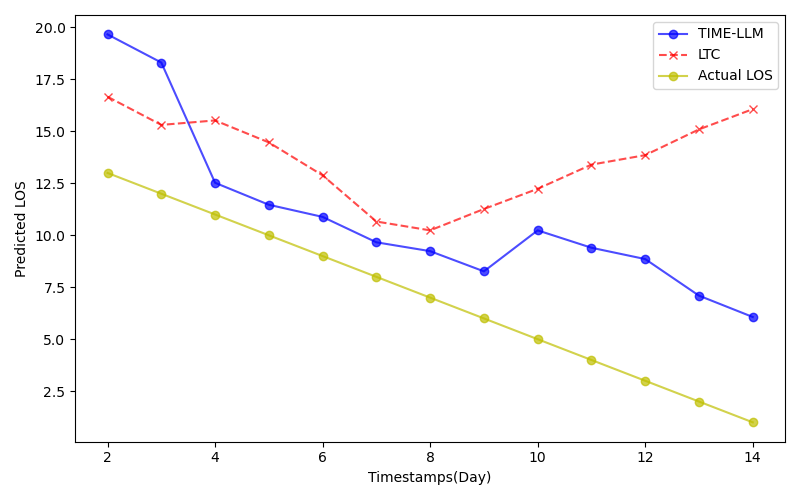}
  \caption{Predictions of real-time LOS for a Deceased patient using autoencoded health vectors and SOI scores: LTC vs. TIME-LLM }
  \label{LTC vs TIME-LLM mortality}
\end{figure}

Figure \ref{LTC vs TIME-LLM} illustrates the comparison of the two most effective time series prediction framework LTC and TIME-LLM for timestamped LOS prediction, specifically from day 2 for a patient who was discharged on the 25th day after admission. A careful analysis of the erroneous predictions, as shown in Figure \ref{LTC vs TIME-LLM mortality}, revealed that mispredictions occurred for certain patients, particularly those who experienced significant complications or expired. Our model predicted these cases as admissions requiring a long stay, which aligns with the health features present in the data. This underscores the effectiveness of nursing notes in reflecting a patient's true condition, suggesting that such information should be integrated separately into the prediction model.

\begin{table}[ht]
\footnotesize
\caption{Comparison of the number of parameters, memory usage, and training time across different models for LOS prediction task.}
\renewcommand{\arraystretch}{1.5}
\begin{tabular}{|p{.8in}|p{.4in}|p{.4in}|p{1in}|} 
\hline
Model & No. of parameters & Memory usage & Time \\
\hline
LSTM & 648K & 34.7 MB CPU & $\approx$ 18 min (training 100 epochs, batch size 16) \\
\hline
ClinicalBERT+LSTM & 111M & 11.8 GB GPU & $\approx$ 1.5 hr (finetuning 20 epochs, batch size 16) \\
\hline
BlueBERT+LSTM & 111M & 12.9 GB GPU & $\approx$ 2 hr (finetuning 20 epochs, batch size 16) \\
\hline
Informer & 1.05M & 1.2 GB GPU & $\approx$ 1.5 hr (training 20 epochs, batch size 16)\\
\hline
TIME-LLM & 7B & 28 GB GPU & $\approx$ 3 hr (training 10 epochs, batch size 16) \\
\hline
LTC & 100K & 1.2 MB CPU & $\approx$ 1 hr (training 100 epochs, batch size 16)\\
\hline
\end{tabular}
  \label{memory-table}

\end{table}

An important observation is that LTC networks are highly resource-efficient, making them well-suited for CPU usage, unlike transformers and large language models (LLMs) that demand significant GPU resources—typically around 20 GB of GPU memory. In contrast, LTC models require less than 1.2 MB of memory while achieving comparable performance, highlighting their minimal memory footprint. As shown in Table \ref{memory-table}, compared to TIME-LLM, Liquid Neural Networks offer notable resource efficiency, utilizing only 100K parameters and 28 neurons in the AutoNCP wiring. While training LLMs is resource- and time-intensive, the LTC model can complete training for 100 epochs in just one hour on the proposed datasets.

\section{Conclusion}
In conclusion, this research introduced Liquid Time-Constant Networks for predicting real-time hospital Length of Stay, effectively capturing complex temporal dynamics in sequential data. The StayLTC model outperformed traditional time series and transformer models, and in some cases, even surpassed large language models in resource-constrained settings. Moving forward, we aim to explore the model's explainability, enabling the identification of key features driving predictions to foster trust among clinical practitioners. However, the framework's performance is contingent on the quality and completeness of unstructured clinical notes, and extracting data from Electronic Health Records requires substantial time for cleaning and interpretation to ensure accurate LOS predictions.



\bibliographystyle{ACM-Reference-Format}
\bibliography{sample-base}










\end{document}